\documentclass[lettersize,journal]{IEEEtran}
\usepackage{amsmath,amsfonts}
\usepackage{algorithmic}
\usepackage{algorithm}
\usepackage{array}
\usepackage[caption=false,font=normalsize,labelfont=sf,textfont=sf]{subfig}
\usepackage{textcomp}
\usepackage{stfloats}
\usepackage{url}
\usepackage{verbatim}
\usepackage{graphicx}
\usepackage{cite}
\usepackage{hyperref}
\usepackage{color}
\usepackage{colortbl}
\usepackage{booktabs}
\usepackage{makecell,multirow,diagbox}
\begin{document}

\title{URoadNet: Dual Sparse Attentive U-Net for Multiscale Road Network Extraction}

\author{Jie Song, \IEEEmembership{Member, IEEE}, Yue Sun, Ziyun Cai, \IEEEmembership{Member, IEEE}, Liang Xiao, \IEEEmembership{Senior Member, IEEE}, \\Yawen Huang, \IEEEmembership{Member, IEEE}, and Yefeng Zheng, \IEEEmembership{Fellow, IEEE}
\vspace{-2em}
\thanks{This work was supported in part by the Frontier Technologies R\&D Program of Jiangsu under grant BF2024070; in part by the China Postdoctoral Science Foundation under Grant 2021M691656; in part by the Natural Universities in Jiangsu Province under Grant 20KJB520005; and in part by the Jiangsu Geological Bureau Research Project under Grant 2023KY11. \emph{(Corresponding author: Liang Xiao.)}}
\thanks{J. Song, Y. Sun, and Z. Cai are with the College of Automation \& College of Artificial Intelligence, Nanjing University of Posts and Telecommunications, Nanjing 210023, China (e-mail: j.song@njupt.edu.cn; yuesun0122@163.com; caiziyun@163.com).}
\thanks{L. Xiao is with the School of Computer Science and Engineering, Nanjing University of Science and Technology, Nanjing 210094, China, also with the Key Laboratory of Intelligent Perception and Systems for High-Dimensional Information of Ministry of Education, Nanjing University of Science and Technology, Nanjing 210094, China, and also with Jiangsu Key Laboratory of Spectral Imaging \& Intelligent Sense, Nanjing University of Science and Technology, Nanjing 210094, China (e-mail: xiaoliang@mail.njust.edu.cn).}
\thanks{Y. Huang and Y. Zheng are with the Jarvis Research Center, Tencent YouTu Lab, Shenzhen, 518057, China (e-mail: bear\_huang@126.com; yefeng.zheng@gmail.com).}}

\markboth{Journal of \LaTeX\ Class Files}%
{Shell \MakeLowercase{\textit{et al.}}: A Sample Article Using IEEEtran.cls for IEEE Journals}


\maketitle

\begin{abstract}
The challenges of road network segmentation demand an algorithm capable of adapting to the sparse and irregular shapes, as well as the diverse context, which often leads traditional encoding-decoding methods and simple Transformer embeddings to failure. We introduce a computationally efficient and powerful framework for elegant road-aware segmentation. Our method, called URoadNet, effectively encodes fine-grained local road connectivity and holistic global topological semantics while decoding multiscale road network information.
URoadNet offers a novel alternative to the U-Net architecture by integrating connectivity attention, which can exploit intra-road interactions across multi-level sampling features with reduced computational complexity. This local interaction serves as valuable prior information for learning global interactions between road networks and the background through another integrality attention mechanism. The two forms of sparse attention are arranged alternatively and complementarily, and trained jointly, resulting in performance improvements without significant increases in computational complexity.
Extensive experiments on various datasets with different resolutions, including Massachusetts, DeepGlobe, SpaceNet, and Large-Scale remote sensing images, demonstrate that URoadNet outperforms state-of-the-art techniques. Our approach represents a significant advancement in the field of road network extraction, providing a computationally feasible solution that achieves high-quality segmentation results.
\end{abstract}
\vspace{-0.1cm}

\section{Introduction}
\IEEEPARstart{R}{oad} structures appear at many scales and in many contexts \cite{c1}, \cite{c2}, \cite{c3}, \cite{c4}, \cite{c5}. They can be one-way streets, two-lane roads in residential areas, cart tracks, bridges, and even highways. As a result, research on them is required in many applications such as navigation, autonomous driving, urban planning, and smart city construction. Many segmentation algorithms learn to convert the raw pixel content from complex road network into more informative geometric and topological descriptions. Since the road structures share multiscale features of being slender and tortuous, they need strong capabilities to model the finer local pixel details and holistic global topological semantics. State-of-the-art techniques rely on variants of U-Net \cite{c6} designed to decode local spatial context at varying levels of complexity \cite{c7}, \cite{c8}, recurrently refined for specific profiles \cite{c9}, \cite{c10}, or attending to all the related features within architectures \cite{c11}, \cite{c12}, \cite{c13}, \cite{c14}. They propagate or model the multi-level road structural features for assisting dense road network prediction process.

Among these methods, the self-attention embeddings tend to outperform the traditional network designs when the road structures become extremely irregular and contaminated by the surrounding covers. Conceptually, U-Net \cite{c6} comprises three scalable parts: an encoding path, a symmetric decoding path, and the skip-connections. Some works aim at improving the encoder \cite{c11}, \cite{c13} or decoder \cite{c12}, \cite{c15} parts, in other words combining self-attention with downsampling or upsampling. This way replaces the standard global attention and naturally enables the learning of intra-road connectivity information, which further boosts the learning of finer local road features. Others focus on the bottleneck \cite{c16} or skip connections \cite{c14}, \cite{c17} between the encoder and decoder and trigger inter-road interactions for global integrality via channel attentions. They introduce multiscale channel-wise information to U-Net while simultaneously reduce the self-attention computational complexity.

Despite their interesting designs and good performance, the prior works have the following issues: 1) They attain various tradeoffs between image resolution and global context: spatial and channel self-attention embeddings suffer either from lack of finer local road perception or loss of variable global structural semantics. {\em Beyond these various variants, can we develop a powerful dual attention embedding that explores holistic interactions from full scales?} 2) Classical segmentation models usually exploit multiscale features, where thin and fragile road structures need to be segmented from high-resolution feature maps. Simply applying Transformer embedding on high-resolution feature maps is not efficient due to the sparsity of the road structures and not practical due to the prohibitive computational cost. {\em Based on dual attention, can we develop an efficient mechanism to attend to sparse tokens from both pixel and semantic perspectives?}

\begin{figure}[!t]
  \centering
  \includegraphics[width=0.98\linewidth]{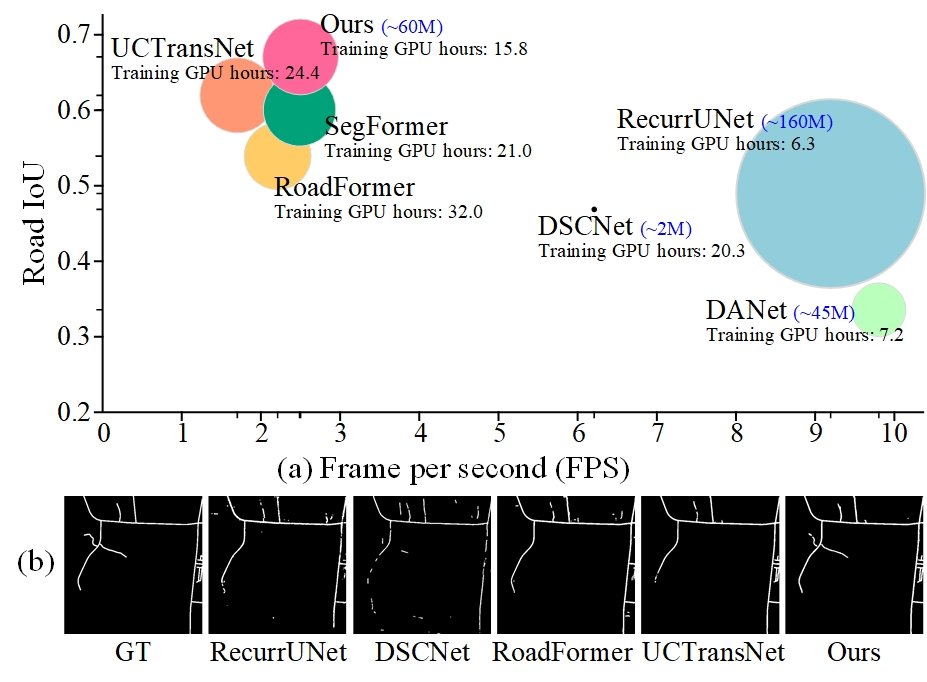}
  \vspace{-0.25cm}
   \caption{\footnotesize{Speed versus accuracy for road segmentation. Each circle depicts the performance of a model in terms of frames-per-second and Road IoU accuracy on the Massachusetts dataset using an NVIDIA GeForce RTX 2080 Ti$^{\circledR}$ GPU. The circle size is proportional to the number of the parameters of the model. We plot the performance of our method in red and show the comparative segmentations at the bottom.}}
  \label{fig1}
\end{figure}

In this paper, we show that these issues can be addressed by reformulating multiscale road feature extraction in terms of a dual sparse attention learning problem. Taking advantage of the dual sparse attention embedding, we present an efficient framework, named URoadNet, for road-aware segmentation. More precisely, we decompose the training into connectivity and integrality attentions to return finer local road details and holistic global structural semantics, each attention being imposed by dependency on the other. In this way, performing pixel-level labeling on their output produces the whole road network contents at once. We will show that, on very irregular roads contaminated by the surrounding covers, e.g., the shadows and occlusions of buildings or vegetation, it outperforms the recent strip convolution-based approach \cite{c18}; the powerful U-Net extensions \cite{c7}, \cite{c10}, \cite{c13}, \cite{c14} developed to improve the performance on the irregular structures; and several other multiscale attention methods \cite{c19}, \cite{c20}. We also evaluate the ability of our method to ease the training with reduced token sizes and sparse attentions. In particular, we propose the connectivity attention to dynamically attend the intra-road positions along the centerline within each local window and the integrality attention to model the semantic inter-dependencies in global spatial dimension. The two types of attention are alternatively arranged, complementary, and jointly trained via interleaved update, offering a complementary alternative to U-Net interaction of information in a computationally efficient manner. Extensive experiments on various datasets, including the Massachusetts roads \cite{c2}, DeepGlobe roads \cite{c3}, SpaceNet roads \cite{c4}, and Large-Scale remote sensing images \cite{c5} demonstrate the computational efficiency of our method, see Fig.~\ref{fig1}.

Our contributions are summarized below:
\begin{itemize}
\renewcommand{\labelitemi}{$\vcenter{\hbox{\tiny$\bullet$}}$}
\item We propose URoadNet, the first multiscale road-aware segmentation framework. Inside URoadNet, connectivity pathway and integrality pathway are tailored in an interleaved way to model intra- and inter-road interaction from full scales, enabling improved performance over the state-of-the-art.
\item We redesign self-attention patterns in both connectivity attention and integrality attention, dubbed dual sparse attention, enabling sparse sampling of pixels (local spatial information) and semantics (global spatial information) and hence reducing quadratic complexity to linear.
\item URoadNet improves three representative Transformer embeding designs in term of Road IoU by 7.5\%/13.8\%/ 5.7\%, 0.4\%/8.1\%/10.3\%, and 10.0\%/17.9\%/16.4\% on Massachusetts, DeepGlobe, and SpaceNet, respectively, with better speed-accuracy tradeoffs. In regard of Large-Scale (LS) road network prediction, URoadNet also achieves top performance with 67.8\%, 77.0\%, 71.0\%, and 69.0\% Road IoU on Massachusetts{\color{red}$^{\rm \bf{LS}}$}, Boston{\color{red}$^{\rm \bf{LS}}$}, Birmingham{\color{red}$^{\rm \bf{LS}}$}, and Shanghai{\color{red}$^{\rm \bf{LS}}$} images.
\end{itemize}

\section{Related work}
\subsection{Non-Self-Attention Variations of U-Net}
Various CNN variants of U-Net have been developed and applicable for road network prediction; for example, UNet++ \cite{c7} developed a nested U-Net architecture, Deep ResUnet \cite{c21}, D-LinkNet \cite{c22}, and MSMDFF-Net \cite{c23} developed a road segmentation algorithm based on residual U-Net, and CasMT \cite{c24}, RoadCorrector \cite{c25}, and OSM-DOER \cite{c49} proposed a multi-branch framework for road network extraction.

Recurrent U-Net (RecurrUNet) \cite{c10} alternatively pioneered an RNN-based design, allowing us to propagate the higher-level semantics through the gated recurrence, and, in conjunction with a progressive refinement on the segmentation mask. Some of the other relevant works on road structure segmentation include RCNN-UNet \cite{c26} for road centerline extraction, SC-RoadDeepNet \cite{c27} for road connectivity enhancement, and PL-WGAN \cite{c28} for urban road network construction.

\subsection{Self-Attention Embeddings of U-Net}
Recently, U-Net segmentors with various self-attention embeddings have seen more rapid progress compared to classical designs. These embeddings include dual self-attention \cite{c14}, \cite{c17}, \cite{c29}, convolutional vision Transformer \cite{c11}, \cite{c30}, \cite{c31}, non-local learning \cite{c12}, \cite{c32}, and deformable attention \cite{c13}, \cite{c33}, \cite{c34}. As a result, NL-LinkNet \cite{c32} beated the D-LinkNet in DeepGlobe Road Extraction Challenge for the first time as a Transformer embedding U-Net segmentation model. Earlier methods resorted to simple bottom-up \cite{c35} and top-down \cite{c12}, \cite{c36} feature integrations. NL-LinkNet \cite{c32} and the following work \cite{c37} proposed to decouple content and positional information to provide better local spatial priors in segmentation. UCTransNet \cite{c14} further designed better formulations of global image-level tokens than previous works. Instead, our method is based on simultaneous modeling of local pixel details and global image semantics, which is stronger and more flexible.

There also exist some embeddings that resort to deformable mechanisms \cite{c38}, \cite{c39}, \cite{c40}. In these methods, a small set of key positions around a reference point are learned from the sampling offsets, thus augmenting the original model with higher flexibility and efficiency. The reference point and sampling strategy could be various. DSCNet \cite{c18} selects the subsequent reference point to be observed in turn for each segment to be processed. RoadFormer \cite{c13} uniformly samples a set of reference points across the road map with a grid partition structure and the following work \cite{c41} employs $1\times1$ convolution on the search space to a local scope of interrelated pixels for the reference. Since most sparse patterns leave the network completely free to learn the geometric changes, they tend to deviate from the true road structures. Instead of free learning, URoadNet introduces road-aware sparse samplings for intra- and inter-road feature encoding-decoding.

\begin{figure}[!t]
  \centering
  \includegraphics[width=0.98\linewidth]{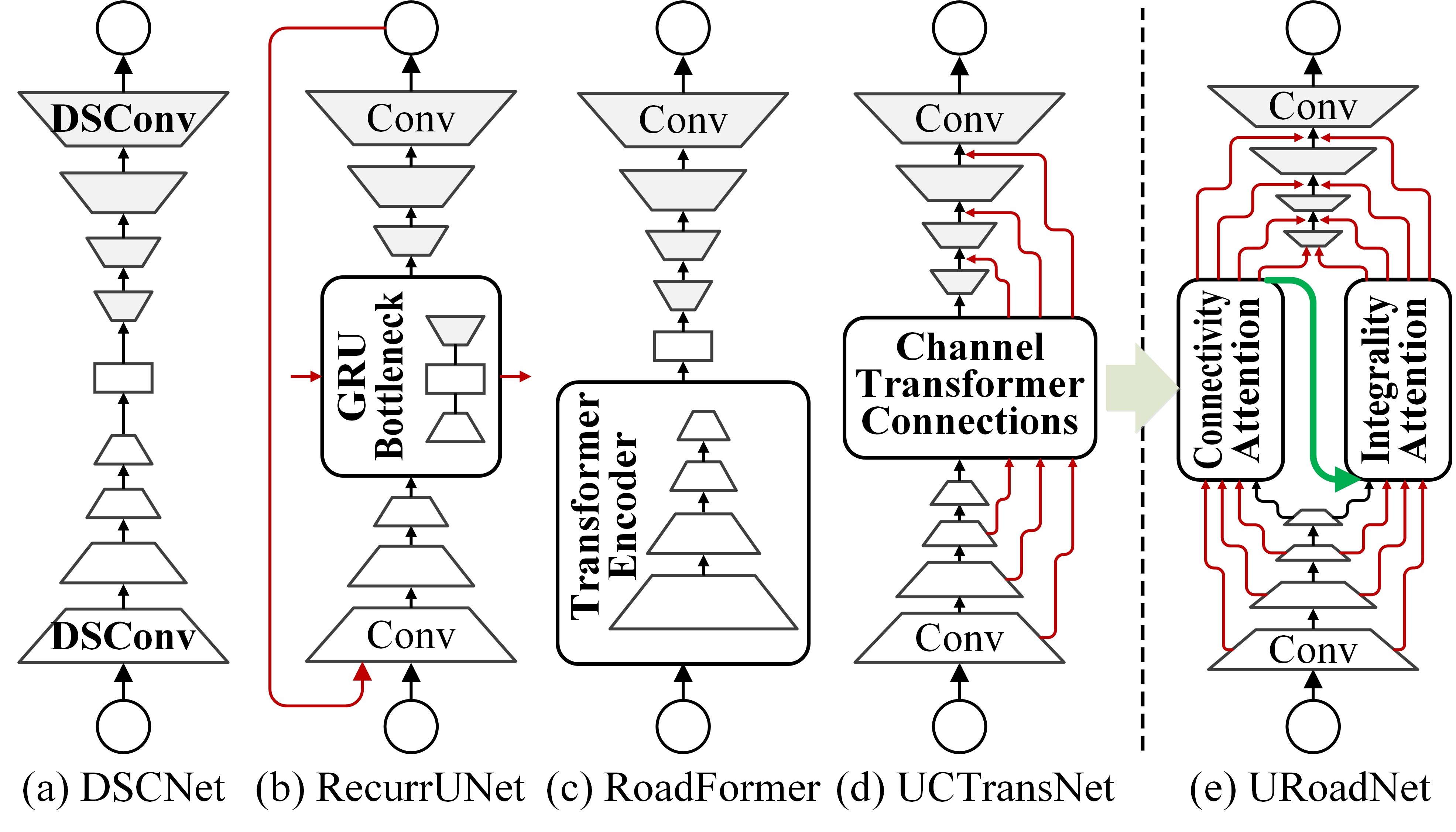}
  \vspace{-0.25cm}
   \caption{\footnotesize{U-Net extensions for road structure segmentation. (a) variant of \cite{c18} based on strip convolution focuses only on finer local characteristics, while (b-d) variants can capture global characteristics by integrating recursions on network layers \cite{c10}, combining self-attention with downsampling \cite{c13}, and replacing skip connections with Transformer embedding \cite{c14}, respectively. By contrast, we propose to decompose and integrate the learning into connectivity and integrality attentions to return both finer details and holistic semantics.}}
  \vspace{-0.1cm}
  \label{fig2}
\end{figure}

\setlength{\tabcolsep}{1pt}
\begin{table}[!t]
  \centering
  \small
  \caption{A Comparative Study of Various U-Net Architectures and the New URoadNet Architecture on the Massachusetts Dataset.
  Here We Use the DRU-ResNet50 as the Version of {\bf Recurr}ent {\bf U}-{\bf Net}. P and R Refers to {\bf P}recision and {\bf R}ecall.}
  \vspace{-0.15cm}
  \begin{tabular}{l|cccccc}
  \toprule[1pt]
  {\bf Architecture} & {\bf Params} & {\bf FLOPs} & {\footnotesize{\bf \makecell[c]{Training \\GPU hours}}} & {\footnotesize{\bf \makecell[c]{Inference\\FPS}}} & {\bf P (\scalebox{0.8}{\%})} & {\bf R (\scalebox{0.8}{\%})}\\
  \midrule[1pt]
  UNet++\cite{c7} & 9.2M & 139.6G & {\bf 5.5} & {\bf 13.6} & 58.6 & 72.5\\
  RecurrUNet\cite{c10} & 167.0M & 478.5G & 6.3 & 9.2 & 61.3 & 69.9\\
  DSCNet\cite{c18} & {\bf 1.6M} & {\bf 39.9G} & 20.3 & 6.2 & {\bf 81.6} & 52.2\\
  \midrule
  RoadFormer\cite{c13} & 59.2M & \cellcolor[rgb]{0.9,0.9,0.85}277.4G & \cellcolor[rgb]{0.9,0.9,0.85}32.0 & \cellcolor[rgb]{0.9,0.9,0.85}2.2 & \cellcolor[rgb]{0.9,0.9,0.85}62.1 & \cellcolor[rgb]{0.9,0.9,0.85}77.5\\
  UCTransNet\cite{c14} & 66.2M & \cellcolor[rgb]{0.9,0.9,0.85}172.0G & \cellcolor[rgb]{0.9,0.9,0.85}24.4 & \cellcolor[rgb]{0.9,0.9,0.85}1.7 & \cellcolor[rgb]{0.9,0.9,0.85}74.2 & \cellcolor[rgb]{0.9,0.9,0.85}79.2\\
  \midrule
  {\bf URoadNet} & 66.9M & \cellcolor[rgb]{0.9,0.9,0.85}99.7G & \cellcolor[rgb]{0.9,0.9,0.85}15.3 & \cellcolor[rgb]{0.9,0.9,0.85}2.5 & \cellcolor[rgb]{0.9,0.9,0.85}79.5 & \cellcolor[rgb]{0.9,0.9,0.85}{\bf 83.0}\\
  \bottomrule
  \end{tabular}
  \vspace{-0.25cm}
  \label{tab1}
\end{table}

\section{Method}
We now introduce our novel multiscale road-aware segmentation architecture. Fig.~\ref{fig2} illustrates how URoadNet evolves from the original U-Net and differs from the most representative segmentation variants. In the following, we first trace these evolutions and differences, motivating an alternative choice for them, and then discuss the technical and implementation details of URoadNet.

\subsection{Motivation Behind the New Architecture}
We have done a comparative study to investigate the performance of various U-Net extensions. To this end, we use the challenging Massachusetts roads dataset. Table~\ref{tab1} summarizes the results including parameters, FLOPS, speed, and accuracy for Massachusetts roads. Our experiments suggest two key findings: 1) Transformer embedding U-Net designs are not necessarily always better, especially in learning finer local road details and controlling the computational complexity; 2) Network variation with recurrent deployment may be feasible for multiscale structure segmentation and much faster, but its model complexity depends on the selection of the backbone and the number of recurrent iterations. While these findings might be attributed to the facts: First, high computational requirements of self-attention block their implementation on high-resolution segmentation. Although deformable \cite{c13} and CTrans \cite{c14} strategies can mitigate the computations with high-resolution inputs, their free learning and global modeling inevitably ignore the intra-road interaction. Second, the earlier attempts rely heavily on the downsampling of the feature maps into restrictive regions, thereby limiting the range of holistic interaction from full scales. Therefore, they need extra processing to obtain the desired results, leading to varied complexity.

Actually, a discussion of deriving dual-path theory in \cite{c19} offers an important inspiration for the answer to these questions, which opens up the possibility of exploiting both local and global visual dependencies from dual attention learning. We explore a more elegant mechanism than that in \cite{c19} to answer questions pertaining, not to independently dense self-attention computations, but to complementarily sparse key sampling for learning multiscale road-aware segmentation. As detailed in Table~\ref{tab1}, the number of parameters of URoadNet is on par with RoadFormer and UCTransNet, but it is faster and obtains 17.4\%/ 5.5\% and 5.3\%/3.8\% better precision/recall, respectively.

\begin{figure*}[t]
  \centering
  \includegraphics[width=0.98\linewidth]{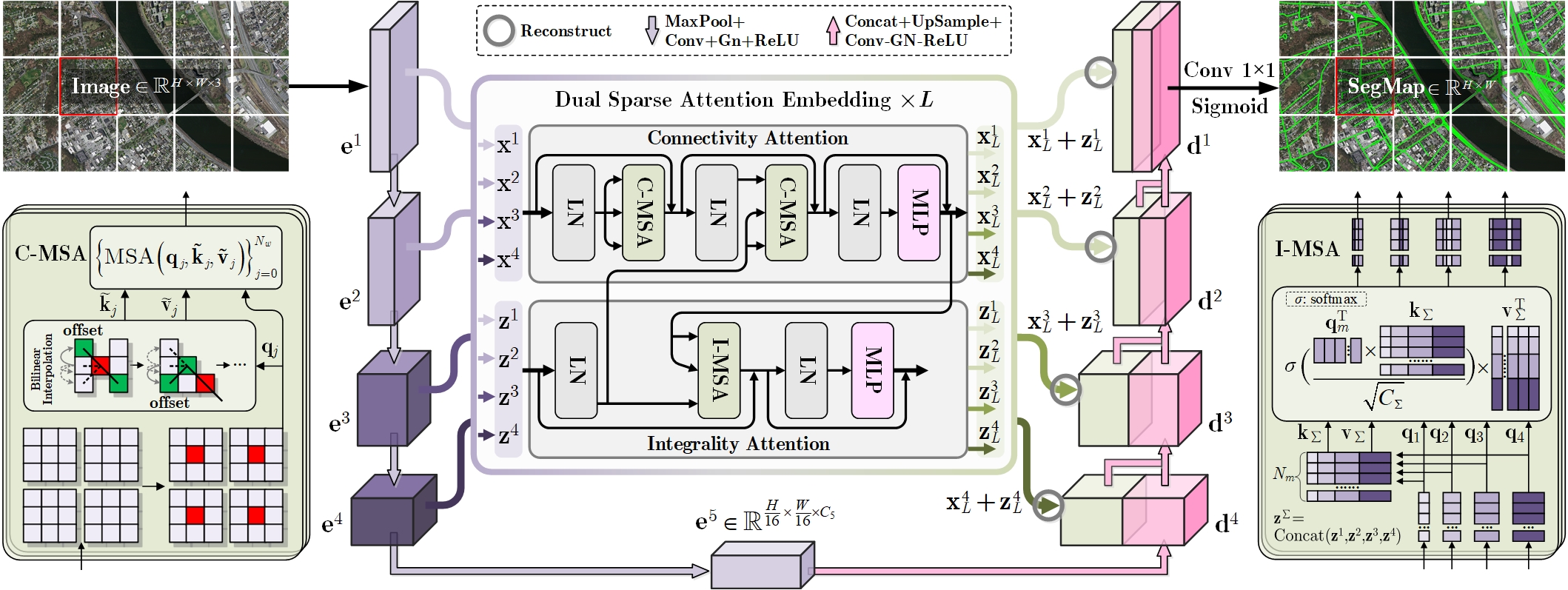}
   \caption{Proposed URoadNet with Dual-SA embedding. We use a convolution layer with sigmoid to compute the final segmentation map. More detail in Section III.}
  \vspace{-0.1cm}
  \label{fig3}
\end{figure*}

\subsection{Model Overview}
Fig.~\ref{fig3} illustrates the detailed schematic. Following the idea from \cite{c19}, URoadNet consists of a U-Net backbone and a dual sparse attention (Dual-SA) embedding. Both the encoder and decoder are comprised of four ``Conv-Gn-ReLU'' stages where the first stage has 64 feature channels, and the channel number doubles after every pooling layer in the encoder. The decoder uses bilinear interpolation for simplicity and group normalization (Gn) \cite{c42} for small batch sizes. The Dual-SA embedding aims to fuse the information across the key road details and key structural semantics at multiple scales. It models sampled features from all stages in the backbone, i.e., ${{\bf{e}}^1}$, ${{\bf{e}}^2}$, ${{\bf{e}}^3}$, and ${{\bf{e}}^4}$, and forwards them through two interactive pathways. The resulting features are fused and then reconstructed through an upsampling operation followed by a convolution layer, and concatenated with the decoder features ${{\bf{d}}^1}$, ${{\bf{d}}^2}$, ${{\bf{d}}^3}$, and ${{\bf{d}}^4}$, respectively, for final high-resolution segmentation.

Our contributions are to decompose the training into 1) local connectivity attention and 2) global integrality attention via the proposed Dual-SA and to integrate these two pathways that complement each other on local and global spatial sampling. The connectivity attention adaptively focuses on the intra-road positions along the centerline within each local window. After that, the integrality attention takes these local connectivities as rich prior in the form of keys/values to refine input feature maps via cross-attention. We divide the Dual-SA into four stages, where a multiscale feature embedding layer is inserted at the beginning of each stage, and stack Dual-SA block in each stage with the feature dimension and resolution kept the same. Below, we first reformulate the general dual attention \cite{c19} in Transformer form, and then discuss our two multiscale road-aware variants, i.e., Dual-SA.

\subsection{Dual Sparse Attention Embedding}
\subsubsection{New Reformulation}
Taking a flattened single-scale feature map ${\bf{x}}_{\ell},{\bf{z}}_{\ell} \in {\mathbb{R}}{^{N \times C}}$ as the input to the pixel and the semantic pathways \cite{c19}, respectively, where $C$ is the number of channels and $N = H \times W$ ($H$/$W$: height/width) is the number of tokens, the ${\ell}$-th dual attention Transformer block can be formulated as
\begin{equation}
\begin{array}{l}
\vspace{0.5ex}
{\kern 10pt}{{{\bf{\bar x}}}_{\ell}} = {\rm{LN}}\left( {{{\bf{x}}_{\ell}}} \right),{{{\bf{\bar z}}}_{\ell}} = {\rm{LN}}\left( {{{\bf{z}}_{\ell}}} \right),\\
\vspace{0.5ex}
{\kern 7.3pt}{{{\bf{x'}}}_{\ell}} = {\rm{MS}}{{\rm{A}}}\left( {{{{\bf{\bar x}}}_{\ell}},{{{\bf{\bar x}}}_{\ell}},{{{\bf{\bar x}}}_{\ell}}} \right) + {{\bf{x}}_{\ell}},\\
\vspace{0.5ex}
{{\bf{x}}_{{\ell} + 1}} = {\rm{MLP}}\left( {{\rm{LN}}\left( {{{{\bf{x'}}}_{\ell}}} \right)} \right) + {{{\bf{x'}}}_{\ell}},\\
\vspace{0.5ex}
{\kern 8.5pt}{{{\bf{z'}}}_{\ell}} = {\rm{MS}}{{\rm{A}}}\left( {{{{\bf{\bar z}}}_{\ell}},{{{\bf{\bar z}}}_{\ell}},{{{\bf{\bar z}}}_{\ell}}} \right) + {{\bf{z}}_{\ell}},\\
\vspace{0.5ex}
{\kern 1.2pt}{{\bf{z}}_{{\ell} + 1}} = {\rm{MLP}}\left( {{\rm{LN}}\left( {{{{\bf{z'}}}_{\ell}}} \right)} \right) + {{{\bf{z'}}}_{\ell}},\\
{\kern 0.5pt}{{\bf{y}}_{{\ell} + 1}} = {{\bf{x}}_{{\ell} + 1}} + {{\bf{z}}_{{\ell} + 1}},
\end{array}
\end{equation}
and
\begin{equation}
\begin{array}{l}
{\kern 9.5pt}{\bf{x'}} = {\rm{Concat}}\left( {{{{\bf{x'}}}^{\left( 1 \right)}},...,{{{\bf{x'}}}^{\left( h \right)}}} \right){{\bf{W}}_o},\\
\vspace{0.5ex}
{{{\bf{x'}}}^{\left( i \right)}} = {\rm{softmax}}\left( {\frac{{{{\bf{q}}^{\left( i \right)}}{{\bf{k}}^{\left( i \right)}}^{\rm{T}}}}{{\sqrt {{d_h}} }}} \right){{\bf{v}}^{\left( i \right)}},i = 1,...,h,\\
\vspace{0.5ex}
{\kern 10.5pt}{\bf{z'}} = {\rm{Concat}}\left( {{{{\bf{z'}}}^{\left( 1 \right)}}^{\rm{T}},...,{{{\bf{z'}}}^{\left( h \right)}}^{\rm{T}}} \right){{\bf{W}}_o},\\
{\kern 1pt}{{{\bf{z'}}}^{\left( i \right)}} = {\rm{softmax}}\left( {\frac{{{{\bf{q}}^{\left( i \right)}}^{\rm{T}}{{\bf{k}}^{\left( i \right)}}}}{{\sqrt {{d_h}} }}} \right){{\bf{v}}^{\left( i \right)}}^{\rm{T}},i = 1,...,h,
\end{array}
\end{equation}
where ${{{\bf{x'}}}^{\left( i \right)}}$ and ${{{\bf{z'}}}^{\left( i \right)}}^{\rm{T}}$ denote the embedding outputs from the $i$-th attention head with dimension ${d_h} = {C \mathord{\left/ {\vphantom {C h}} \right.
 \kern-\nulldelimiterspace} h}$ of different pathways, ${{\bf{q}}^{\left( i \right)}} \!=\! {{\bf{\bar a}}^{\left( i \right)}}{\bf{W}}_q^{\left( i \right)},{{\bf{k}}^{\left( i \right)}} \!=\! {{\bf{\bar a}}^{\left( i \right)}}{\bf{W}}_k^{\left( i \right)},{{\bf{v}}^{\left( i \right)}} \!=\! {{\bf{\bar a}}^{\left( i \right)}}{\bf{W}}_v^{\left( i \right)} \in {\mathbb{R}}{^{N \times d_h}}\left( {{\bf{\bar a}} = {\bf{\bar x}}{\kern 2pt} {\rm{or}}{\kern 2pt} {\bf{\bar z}}} \right)$ represent query, key, and value embeddings in the multi-head self-attention (MSA) module, respectively, and ${\bf{W}}_q^{\left( i \right)},{\bf{W}}_k^{\left( i \right)},{\bf{W}}_v^{\left( i \right)},{{\bf{W}}_o} \in {\mathbb{R}}{^{C \times C}}$ are the projection weights. The dual attention Transformer block first encodes different types of queries via two independent self-attentions, then normalizes their outputs, followed by the same multi-layer perceptron (MLP), and finally fuses two pathways via an element-wise sum.

\subsubsection{Interleaved Token Update}
According to (1), the bottleneck towards an effective dual attention is the independent dense self-attention computations of two pathways. Considering each transposed embedding output contains an abstract representation of the entire road image, e.g., ${{{\bf{z'}}}^{\left( i \right)}}^{\rm{T}}$, such design will inevitably hinder the holistic interactions between low- and high-level feature tokens. Existing techniques, notably DaViT \cite{c43} and CrossViT \cite{c44}, try to address the challenge of interaction. The simple cascading technique of the former and the interaction with a single compressed {\texttt {CLS}} token of the latter result in severe information loss. In the meantime, the observations in \cite{c44} have revealed that using output tokens from different pathways as keys/values to exchange information with each other can benefit the performance. Therefore, we opt for a better solution with interleaved updates for tokens from different pathways.

Specifically, given the input features ${\bf{x}}_{\ell}$ and ${\bf{z}}_{\ell}$ of the ${\ell}$-th Dual-SA block, the connectivity pathway treats the initial integrality features ${\bf{z}}_{\ell}$ as keys/values to interact with the refined connectivity queries ${{{\bf{x'}}}_{\ell}}$ via cross-attention. Formally, we have
\begin{equation}
\begin{array}{l}
\vspace{0.5ex}
{\kern 10pt}{{{\bf{\bar x}}}_{\ell}} = {\rm{LN}}\left( {{{\bf{x}}_{\ell}}} \right),{{{\bf{\bar z}}}_{\ell}} = {\rm{LN}}\left( {{{\bf{z}}_{\ell}}} \right),\\
\vspace{0.5ex}
{\kern 7.2pt}{{{\bf{x'}}}_{\ell}} = {\rm{C\mbox{-}MSA}}\left( {{{{\bf{\bar x}}}_{\ell}},{{{\bf{\bar x}}}_{\ell}},{{{\bf{\bar x}}}_{\ell}}} \right) + {{\bf{x}}_{\ell}},\\
\vspace{0.5ex}
{\kern 10pt}{{{\bf{\hat x}}}_{\ell}} = {\rm{C\mbox{-}MSA}}\left( {{\rm{LN}}\left( {{{{\bf{x'}}}_{\ell}}} \right),{{{\bf{\bar z}}}_{\ell}},{{{\bf{\bar z}}}_{\ell}}} \right) + {{{\bf{x'}}}_{\ell}},\\
{{\bf{x}}_{{\ell} + 1}} = {\rm{MLP}}\left( {{\rm{LN}}\left( {{{{\bf{\hat x}}}}_{\ell}} \right)} \right) + {{{\bf{\hat x}}}_{\ell}},
\end{array}
\end{equation}
where ${\rm{C\mbox{-}MSA}}$ denotes our connectivity MSA, which will be discussed in the following section. Since the connectivity tokens ${{\bf{x}}_{{\ell} + 1}}$ already learn finer local pixel details in its pathway, interacting with the integrality tokens helps to boost semantic discriminability. To do so, the integrality pathway takes ${{\bf{x}}_{{\ell} + 1}}$ as prior information of local connectivity in the form of keys/values to refine initial integrality features ${\bf{z}}_{\ell}$ via cross-attention. Formally, this process can be described as
\begin{equation}
\begin{array}{l}
\vspace{0.5ex}
{{{\bf{\bar x}}}_{{\ell} + 1}} = {\rm{LN}}\left( {{{\bf{x}}_{{\ell} + 1}}} \right),{{{\bf{\bar z}}}_{\ell}} = {\rm{LN}}\left( {{{\bf{z}}_{\ell}}} \right),\\
\vspace{0.5ex}
{\kern 8.3pt}{{{\bf{z'}}}_{\ell}} = {\rm{I\mbox{-}MSA}}\left( {{{{\bf{\bar z}}}_{\ell}},{{{\bf{\bar x}}}_{{\ell} + 1}},{{{\bf{\bar x}}}_{{\ell} + 1}}} \right) + {{\bf{z}}_l},\\
{\kern 1pt}{{\bf{z}}_{{\ell} + 1}} = {\rm{MLP}}\left( {{\rm{LN}}\left( {{{{\bf{z'}}}_{\ell}}} \right)} \right) + {{{\bf{z'}}}_{\ell}},
\end{array}
\end{equation}
where ${\rm{I\mbox{-}MSA}}$ denotes the proposed integrality MSA.

\begin{figure}[!t]
  \centering
  \includegraphics[width=0.98\linewidth]{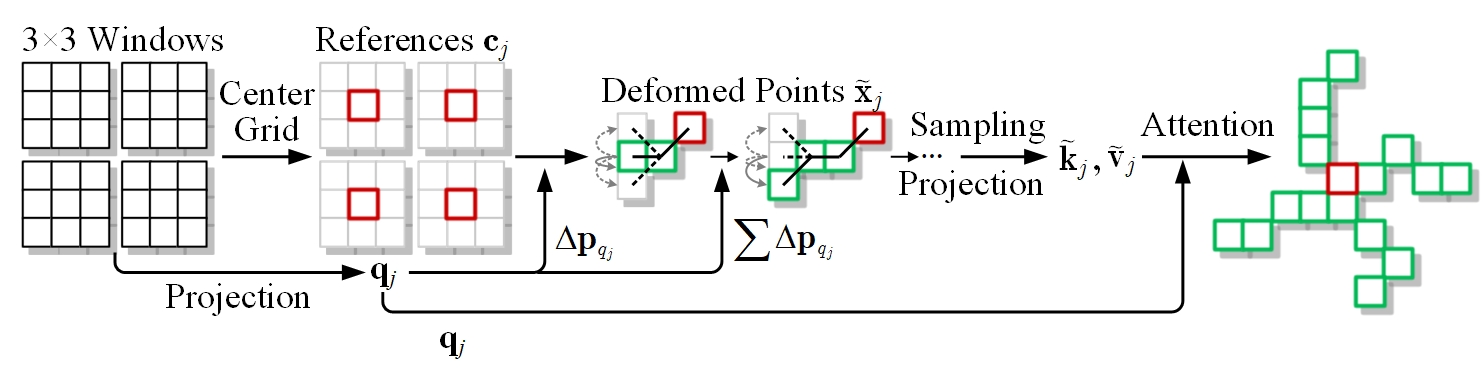}
  \vspace{-0.25cm}
   \caption{\footnotesize{Illustration of our connectivity self-attention mechanism.}}
  \label{fig4}
\end{figure}

Note that both ${{\bf{x}}_{{\ell} + 1}}$ and ${{\bf{z}}_{{\ell} + 1}}$ act as the enhanced queries for next Dual-SA block. Considering the gradients are back propagated through both pathways, Dual-SA block is thus able to simultaneously compensate the local and global information loss.

\subsubsection{Connectivity Self-Attention for Road Awareness}
From our motivation, the bottleneck towards an efficient dual attention is the excessive connectivity features, most of which are not informative for sparse road structures. Moreover, considering $N$ can be very large, e.g., $512 \times 512$, the computation complexity for connectivity pathway above will be very high with numerous query and key elements. Thus a data-dependent sparse self-attention is required to flexibly encode connectivity queries, leading to deformable mechanism. Inspired by \cite{c38}, we introduce deformation offsets $\Delta$. However, if we directly apply the same free learning mechanism in the self-attention, the perceptual field tends to stray outside the target. To better fit the road structure, we propose an iterative method to model the relations among connectivity tokens in turn under the guidance of local centerlines in the feature maps. These focused centerlines are determined by deformed sampling points which are learned from the queries by accumulating sums over offsets within each local window. We adopt bilinear interpolation to sample features, and then get the deformed keys/values from them.

Formally, given the input feature map ${{\bf{e}}^m} \in {{\mathbb{R}}^{H \times W \times C}}$, $m = 1,2,3,4$ (flattened to ${{\bf{x}}^m}$ later), ${N_w}$ center grids ${\bf{c}} \in {{\mathbb{R}}^2}$ of local non-overlapping $3 \times 3$ windows are selected as the initial references, where ${N_w} = {{HW} \mathord{\left/
 {\vphantom {{HW} {{3^2}}}} \right.
 \kern-\nulldelimiterspace} {{3^2}}}$. The connectivity self-attention starts from the center grid ${\bf{c}}$, augments each deformed sampling position within a window with a local offset computed from the position of the previous grid, and is calculated as
\begin{equation}
\begin{array}{l}
\vspace{0.5ex}
{\rm{C\mbox{-}MSA}}\left( {{\bf{q}},{\bf{k}},{\bf{v}}} \right) = \left\{ {{\rm{MSA}}\left( {{{\bf{q}}_j},{{{\bf{\tilde k}}}_j},{{{\bf{\tilde v}}}_j}} \right)} \right\}_{j = 0}^{{N_w}},\\
\vspace{0.5ex}
{\kern 59pt}{{\bf{q}}_j} = {{\bf{x}}_j}{{\bf{W}}_{{q_j}}},{{{\bf{\tilde k}}}_j} = {{{\bf{\tilde x}}}_j}{{\bf{W}}_{{{\tilde k}}_j}},{{{\bf{\tilde v}}}_j} = {{{\bf{\tilde x}}}_j}{{\bf{W}}_{{{\tilde v}}_j}},\\
{\kern 59pt}{{{\bf{\tilde x}}}_j} = {{\bf{x}}_j}\left( {{{\bf{c}}_j} \pm \sum\nolimits_t {\Delta {{\bf{p}}_{{q_j}}}} } \right),
\end{array}
\end{equation}
where ${{\bf{q}}_j},{{{\bf{\tilde k}}}_j},{{{\bf{\tilde v}}}_j}$ are query, deformed key, and deformed value embeddings of the window, respectively; ${\Delta {{\bf{p}}_{{q_j}}}}$ denotes the deformation offset, which is obtained via linear projection over the query ${{\bf{q}}_j}$; and $t \!=\! \left\{ {0,1,2,3,4} \right\}$. So, from a center grid ${\bf{c}}$, two deformed points moving in the opposite direction are generated. Note the superscript $m$ is omitted for simplicity. More details of C-MSA are shown in Fig.~\ref{fig4}.

\begin{figure}[!t]
  \centering
  \includegraphics[width=0.95\linewidth]{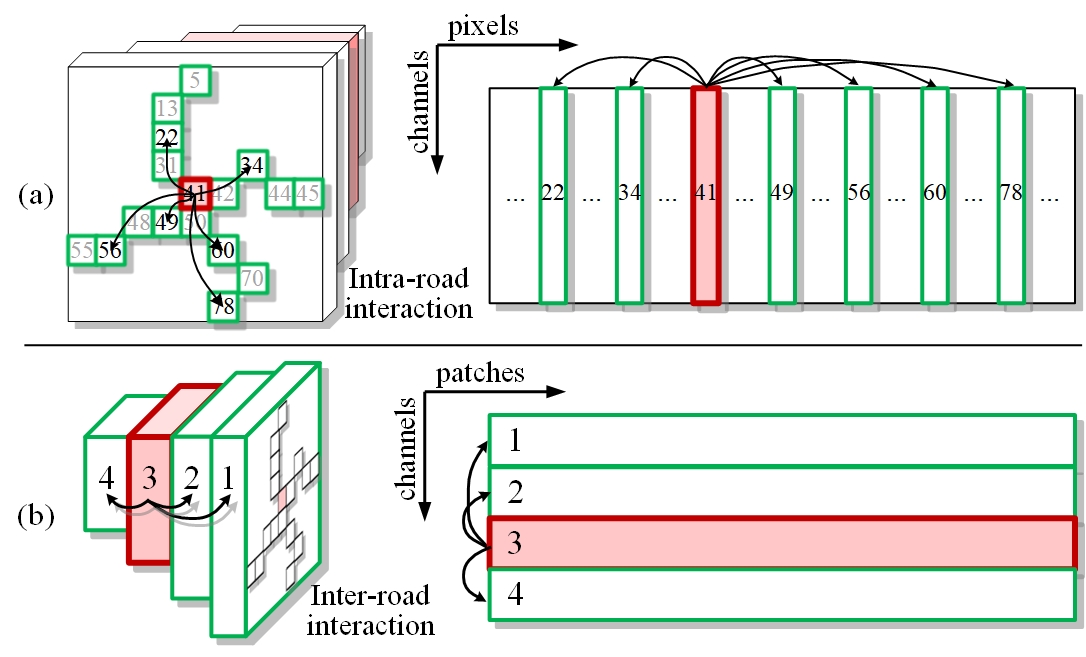}
  \vspace{-0.25cm}
   \caption{\footnotesize{Comparative interpretation of the proposed connectivity self-attention (a) and integrality self-attention (b).}}
  \label{fig5}
\end{figure}

\subsubsection{Integrality Self-Attention for Multiscale Interaction}
Most modern semantic segmentation frameworks benefit from multiscale feature maps \cite{c45}. To further introduce the scale information into the interactions, here we revisit self-attention from another perspective and propose integrality self-attention (see Fig.~\ref{fig5}). Let $\left\{ {{{\bf{z}}^m}} \right\}_{m = 1}^M\left( {M \!=\! 4} \right)$ be the flattened multiscale feature maps from the outputs of the encoder, where ${{\bf{z}}^m} \!\in\! {{\mathbb{R}}^{\frac{{HW}}{{{m^2}}} \times {C_m}}}$. The integrality self-attention linearly projects the feature maps of different scales to obtain the queries ${{\bf{q}}_m} \in {{\mathbb{R}}^{{N_m} \times {C_m}}}$ and then concatenates all scales, i.e., ${{\bf{z}}^\Sigma } \!=\! {\rm{Concat}}\left( {{{\bf{z}}^1},{{\bf{z}}^2},{{\bf{z}}^3},{{\bf{z}}^4}} \right)$, as the keys ${{\bf{k}}_\Sigma }$ / values ${{\bf{v}}_\Sigma } \in {{\mathbb{R}}^{{N_m} \times {C_\Sigma }}}$, where ${N_m} \!=\! \frac{{HW}}{{{m^2}{P_m}^2}}$ is the number of patches and ${P_m}$ corresponds to the resolution of each patch, which is calculated as
\begin{equation}
\begin{array}{l}
\vspace{0.5ex}
{\rm{I\mbox{-}MSA}}\left( {{\bf{q}},{\bf{k}},{\bf{v}}} \right) = {\rm{MSA}}\left( {{{\bf{q}}_m},{{\bf{k}}_\Sigma },{{\bf{v}}_\Sigma }} \right),\\
{\kern 53.9pt}{{\bf{q}}_m} \!=\! {{\bf{z}}^m}{{\bf{W}}_{{q_m}}},{{\bf{k}}_\Sigma } \!=\! {{\bf{z}}^\Sigma }{{\bf{W}}_{{k_\Sigma }}},{{\bf{v}}_\Sigma } \!=\! {{\bf{z}}^\Sigma }{{\bf{W}}_{{v_\Sigma }}}.
\end{array}
\end{equation}
Note compared to ${\rm{C\mbox{-}MSA}}$ that produces an attention map with shape $9 \times 9$, the ${\rm{I\mbox{-}MSA}}$'s attention map has a dimension of ${C_m} \times {C_\Sigma }$, which means the number of patches at each scale in ${\rm{I\mbox{-}MSA}}$ need to be same. Therefore, we use a smaller patch size for higher-level feature maps while a larger patch size for lower-level maps and set ${P_m} = \left\{ {P,\frac{P}{2},\frac{P}{4},\frac{P}{8}} \right\}$. Moreover, a scaling factor $\frac{1}{{\sqrt {{C_\Sigma }} }}$ is computed along the channel dimension, rather than the spatial one. As mentioned earlier, the transposed embedding output ${{\bf{z'}}}$ abstracts the representation of the entire image, making its interactions with global spatial information in linear spatial-wise complexity.

\subsection{Comparison with other Efficient Dual Attention}
We replace the vanilla MSA with our ${\rm{C\mbox{-}MSA}}$ and ${\rm{I\mbox{-}MSA}}$ in (3) and (4), respectively, and integrate them in an interleaved way via cross-attention to build a Dual-SA. Another efficient way is to stack window-channel pairwise self-attention layers, like DaViT, or use a {\texttt {CLS}} to interact with tokens from the other branch, like CrossViT. However, they have two drawbacks: 1) The resulting tokens via simple cascading or compressed {\texttt {CLS}} may not be optimal due to limited interaction. 2) It is hard to generalize across multiscale segmentation tasks because of a lack of efficient processing of high-resolution images.

\begin{figure*}[!t]
  \centering
  \includegraphics[width=0.95\linewidth]{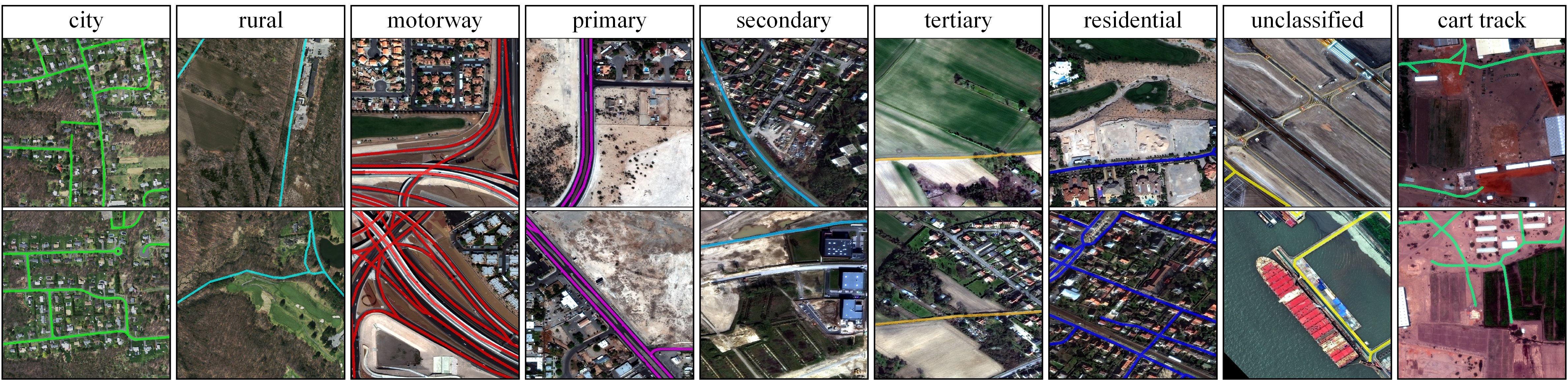}
  \vspace{-0.25cm}
   \caption{\footnotesize{Example roads of different types and areas overlaid with predictive segmentations shown as colored curves.}}
  \label{fig6}
\end{figure*}

\subsection{Road Network Prediction Problem}
In all our experiments, we use the datasets from different satellites, countries, and geographic regions, containing more than 10,000 road images, for training and testing. To ensure richness of road appearances, we cover rural and urban areas, unpaved, paved, and dirt paths. The training set contains tens of thousands of roads that have been annotated over a wide range of scales and contexts. Many roads exhibit intra-road inconsistency while the corresponding images comprises topology-complex road networks and background interference with various surrounding covers, which increase the difficulty of this segmentation task.

We formulate the segmentation problem of diverse roads in the training set as a URoadNet prediction problem. We build a multiscale road-aware predictor with a Dual-SA embedding (Section III-C) to recover fine-grained local and holistic global details. The embedding performs the connectivity attention, integrality attention, and connectivity-integrality interactions by proceeding with an interleaved token update on the sampled road features from full scales. This not only tackled the aforementioned challenges but saved the computational complexity. After applying a single $1 \times 1$ convolution layer and sigmoid function to the final concatenated feature maps of the decoder, we build the road segmentation maps observed in Fig.~\ref{fig6}.

\section{Experiments}
In this section, we first provide detailed instructions on the datasets and the metrics used to test URoadNet. Then, we discuss our results and benchmark state-of-the-art segmentation algorithms based on their properties and public implementations. We then describe our evaluation methodology. Finally, we apply URoadNet to the problem of road network extraction in Large-Scale remote sensing images.

\subsection{Datasets and Metrics}
We conduct experiments on the Massachusetts roads, DeepGlobe roads, SpaceNet roads, and Large-Scale remote sensing images.

\subsubsection{Massachusetts Roads~\cite{c2} [1 m/pixel]}
contains 1,108 training, 14 validation, and 49 testing images with high quality annotations of road and background class, covering more than 2,600 km$^2$ of Massachusett and a wide spectrum of scenarios, including rural, suburban, and urban regions. The size of each image is $1500 \times 1500$.

\subsubsection{DeepGlobe Roads~\cite{c3} [0.5 m/pixel]}
is created as the benchmark dataset for the 2018 Satellite Image Understanding Challenge. This dataset contains data with pixel-level annotations from India, Indonesia, and Thailand. It is randomly divided into 4,891 images as training, 190 as validation, and 1145 as testing. The size of each image is $1024 \times 1024$.

\subsubsection{SpaceNet Roads~\cite{c4} [0.3 m/pixel]}
annotates 2,549 images from four different cities: Shanghai, Paris, Khartoum, and Las Vegas for training and uses the remaining 928 for testing. The ground truth is given in the form of a line-string that indicates the centerline of unpaved, paved, and dirt roads. To benchmark all the methods, road labels are firstly generated from the original GeoJSON files using python package. The size of each image is $1300 \times 1300$.

All the training images of the above three datasets are empirically resized to $512 \times 512$ pixels for efficiency and fair comparison.

\subsubsection{Large-Scale Roads~\cite{c5}}
is collected from Google Earth and accurately labeled for the evaluation. This dataset includes Massachusetts{\color{red}$^{\rm \bf{LS}}$} [1 m/pixel], Boston{\color{red}$^{\rm \bf{LS}}$} [0.44 m/pixel], Birmingham{\color{red}$^{\rm \bf{LS}}$} [0.36 m/pixel], and Shanghai{\color{red}$^{\rm \bf{LS}}$} [0.51 m/pixel] imagery with a pixel resolution of $14116 \times 16273$, $23104 \times 23552$, $22272 \times 22464$, and $16768 \times 16640$, respectively.

\subsubsection{Metrics}
As is generally done in evaluating road segmentation methods \cite{c46}, \cite{c47}, we adopt five local per pixel metrics: F1 score (F1), intersection over union (IoU), precision (P), recall (R), and overall accuracy (OA). The F1 measure is the harmonic mean of the precision and recall at the pixel level that is equivalent to the Dice coefficient.

For the problem of the road network extraction, it is also interesting to examine the global behaviour of a given model as the length of the road path is varied. To this end, we use the overlap (OV) measure that is similar to \cite{c48} and propose the average distance (AD) between the predicted path and the ground truth path. The OV measure is defined as
\begin{equation}
{\rm{OV}} = \frac{{{\rm{TPR + TPM}}}}{{{\rm{TPR + TPM + FN + FP}}}}
\end{equation}
where TPM and TPR are the numbers of true positives in the computed path and the reference path, respectively, and FN and FP are the numbers of false negatives and false positives. Let \emph{PP} and \emph{GP} represent the predicted path and the ground truth path, respectively. The AD measure between \emph{PP} and \emph{GP} is computed as follows
\begin{equation}
{\rm{AD}} = \frac{{{\rm{1}} }}{{\rm{2}}} \left( {\frac{{\sum\nolimits_{x \in PP} { d} \left( {x, GP} \right)}}{{\left| { PP } \right|}} + \frac{{\sum\nolimits_{u \in GP} { d} \left( {u, PP} \right) }}{{\left| { GP } \right|}}} \right)
\end{equation}

\begin{figure*}[!t]
  \centering
  \includegraphics[width=0.98\linewidth]{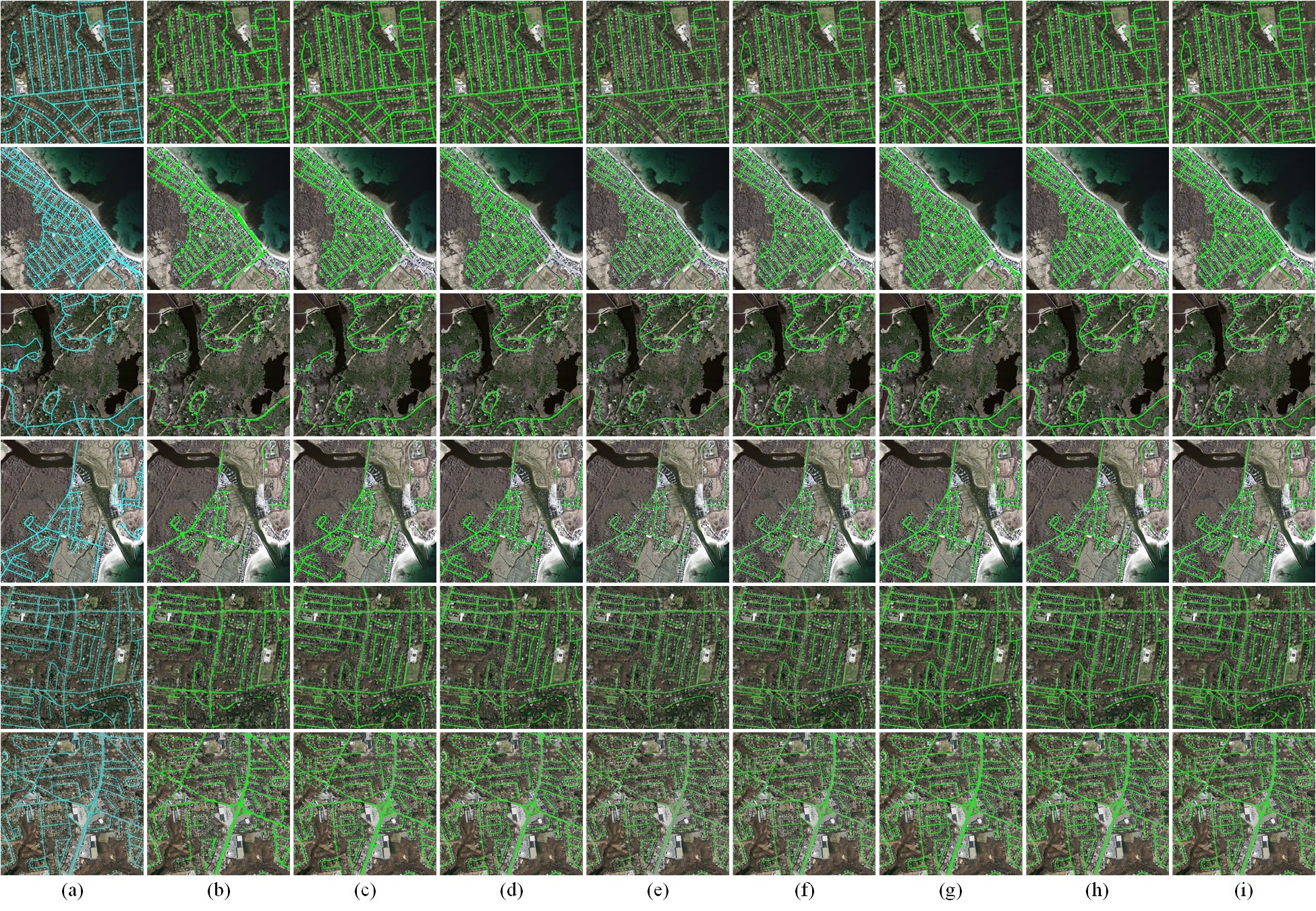}
  \vspace{-0.25cm}
   \caption{Visual results on the test images of the Massachusetts roads. Ground truth is overlaid on the orignal images in (a), followed by segmentations generated by (b) DANet, (c) UNet++, (d) RecurrUNet, (e) DSCNet, (f) SegFormer, (g) RoadFormer, (h) UCTransNet, and (i) URoadNet.}
  \label{fig7}
\end{figure*}

\subsection{Implementation Details}
To ensure convenient and fair comparisons of the performance differences among different algorithms, we dedicate a significant amount of time to debugging all the methods on the PyTorch platform. All the networks share a common training and testing framework and are implemented in PyTorch 1.8.1 on a server with an NVIDIA A40 GPU, an Intel$^{\circledR}$ Xeon$^{\circledR}$ Platinum 8358P (2.60GHz) CPU, and 80G RAM. We utilize Adam with an initial learning rate of 0.0001 to optimize the networks and decrease the learning rate by half when the loss on the validation set has not dropped by 10 epochs. We train all models for 200 epochs with a batch size of two and use the combined cross entropy loss and Dice loss as our loss function.

\setlength{\tabcolsep}{3.5pt}
\begin{table}
  \centering
  \small
  \caption{Ablation Studies of the Proposed C-MSA, I-MSA, and Dual-SA in URoadNet on the Massachusetts and SpaceNet Datasets.}
  \vspace{-0.15cm}
  \begin{tabular}{l|cccc}
  \toprule[1pt]
  \multirow{2}*{} & \multicolumn{2}{c}{\texttt{Massachusetts}} & \multicolumn{2}{c}{\texttt{SpaceNet}} \\
  \cmidrule(r){2-3} \cmidrule(r){4-5}
        & {\bf F1(\scalebox{0.8}{\%})} & {\bf IoU(\scalebox{0.8}{\%})} & {\bf F1(\scalebox{0.8}{\%})} & {\bf IoU(\scalebox{0.8}{\%})} \\
  \midrule[1pt]
  Baseline (U-Net) & 62.22 & 45.75 & 60.70 & 43.99 \\
  + C-MSA & 67.87 & 51.56 & 70.51 & 55.14 \\
  + I-MSA & 71.72 & 57.61 & 75.17 & 62.21 \\
  + Dual-SA & {\bf 80.59} & {\bf 67.61} & {\bf 81.19} & {\bf 69.68} \\
  \bottomrule
  \end{tabular}
  \vspace{-0.25cm}
  \label{tab2}
\end{table}

\setlength{\tabcolsep}{1.2pt}
\begin{table}
\renewcommand\arraystretch{1.5}
  \centering
  \footnotesize
  \caption{Results of Models with Different Label Rate Settings on Massachusetts Dataset.}
  \vspace{-0.15cm}
  \begin{tabular}{l|ccc|ccc|ccc|ccc}
  \toprule[1pt]
  \multirow{2}*{} & \multicolumn{3}{c}{URoadNet} & \multicolumn{3}{|c}{UNet++\cite{c7}} & \multicolumn{3}{|c}{RecurrUNet\cite{c10}} & \multicolumn{3}{|c}{SegFormer\cite{c20}} \\
  \cmidrule(r){2-4} \cmidrule(r){5-7} \cmidrule(r){8-10} \cmidrule(r){11-13}
        & {\bf 100\scalebox{0.8}{\%}} & {\bf 75\scalebox{0.8}{\%}} & {\bf 50\scalebox{0.8}{\%}} & {\bf 100\scalebox{0.8}{\%}}
        & {\bf 75\scalebox{0.8}{\%}}  & {\bf 50\scalebox{0.8}{\%}} & {\bf 100\scalebox{0.8}{\%}} & {\bf 75\scalebox{0.8}{\%}}
        & {\bf 50\scalebox{0.8}{\%}} & {\bf 100\scalebox{0.8}{\%}}  & {\bf 75\scalebox{0.8}{\%}} & {\bf 50\scalebox{0.8}{\%}} \\
  \midrule[1pt]
  F1(\%) & \cellcolor[rgb]{0.9,0.9,0.85}80.6 & \cellcolor[rgb]{0.9,0.9,0.85}79.5 & \cellcolor[rgb]{0.9,0.9,0.85}75.1 & 64.4 & 60.3 & 57.3 & 64.9 & 59.4 & 57.1 & 74.8 & 68.3 & 66.1 \\
  IoU(\%) & \cellcolor[rgb]{0.9,0.9,0.85}67.6 & \cellcolor[rgb]{0.9,0.9,0.85}66.1 & \cellcolor[rgb]{0.9,0.9,0.85}60.3 & 47.7 & 43.4 & 40.5 & 48.3 & 42.5 & 40.2 & 60.1 & 52.0 & 49.8 \\
  \bottomrule
  \end{tabular}
  \vspace{-0.25cm}
  \label{tab3}
\end{table}

\begin{figure*}[!t]
  \centering
  \includegraphics[width=0.98\linewidth]{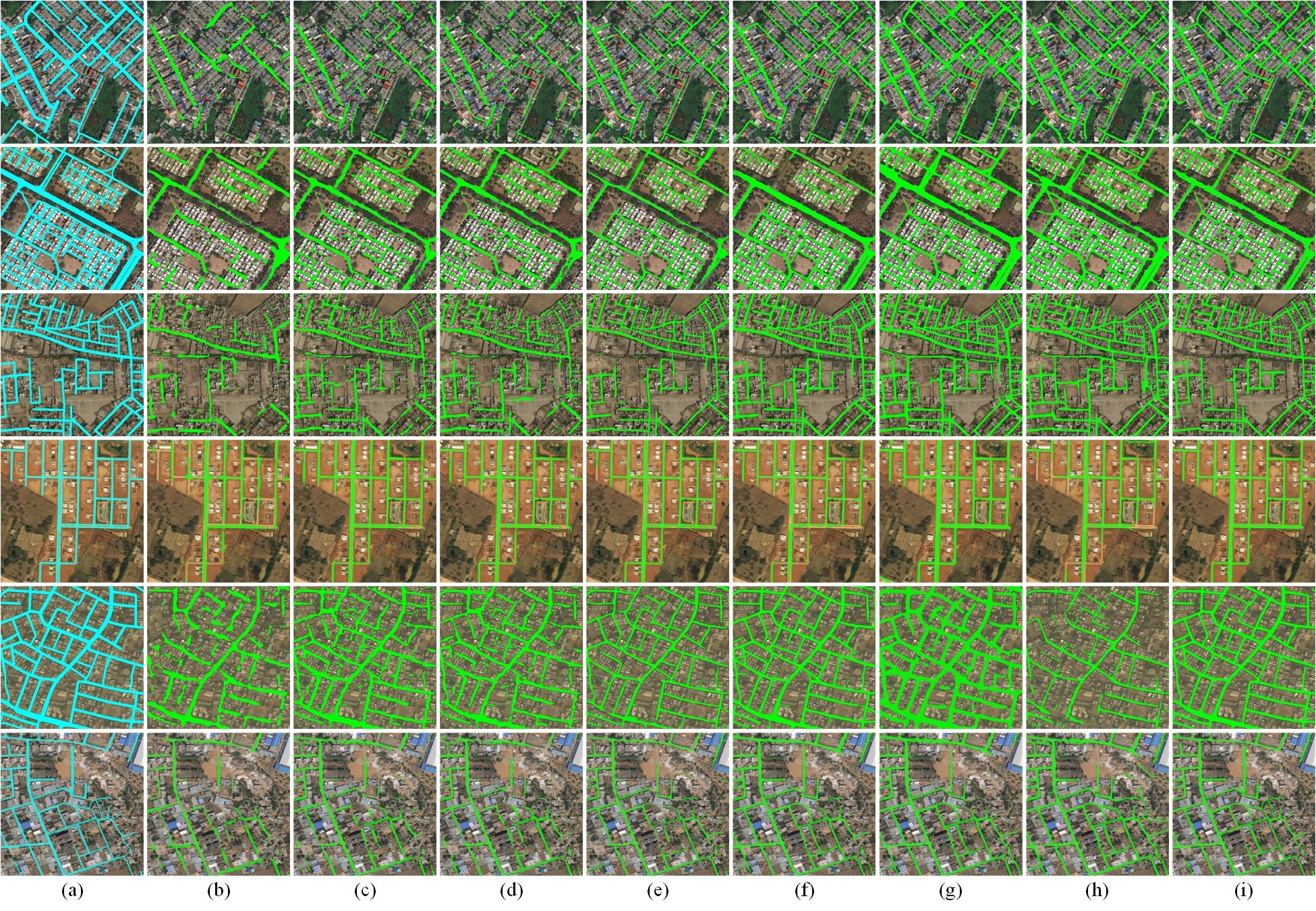}
  \vspace{-0.35cm}
   \caption{Visual results on the test images of the DeepGlobe roads. Ground truth is overlaid on the orignal images in (a), followed by segmentations generated by (b) DANet, (c) UNet++, (d) RecurrUNet, (e) DSCNet, (f) SegFormer, (g) RoadFormer, (h) UCTransNet, and (i) URoadNet.}
  \vspace{-0.25cm}
  \label{fig8}
\end{figure*}

\subsection{Ablation Studies}
\subsubsection{Model}
We first conduct ablation studies on the Massachusetts and SpaceNet datasets which could comprehensively evaluate the ability to handle various types of the roads. The three proposed attention mechanisms inside Dual-SA are evaluated, including our connectivity self-attention (C-MSA), integrality self-attention (I-MSA), and cross-attention from interleaved token update. In these experiments, Dual-SA is trained end-to-end for 4 stages with multi-scale data augmentation. Table~\ref{tab2} summarizes the results. As shown, ``Baseline+Dual-SA'' is generally better than the other ``Baseline+'' on both datasets, which indicates the effectiveness of our unique decomposition and integration. Specifically, C-MSA encodes local road connectivity and I-MSA learns global road network topology. Then they are updated in an interleaved way to attend to the true multiscale paths. The model in the second row keeps the backbone as U-Net and models sampled features from all stages only with C-MSA, which yields 7.73 mF1 and 8.48 mIoU improvements. When exploring the influence of the integrality attention, we add I-MSA before the decoding. Instead of performing attention on pixel-level or patch-level, we apply an attention mechanism on the transpose of patch-level tokens, which compensates the information loss on global feature compression. The comparison between the second and the third row shows 4.26 and 5.87 improvements in terms of mF1 and mIoU. Finally, the interleaved token update is embedded in C-MSA and I-MSA, which achieves 7.45 and 8.74 improvements in terms of mF1 and mIoU.

\begin{figure*}[!t]
  \centering
  \includegraphics[width=0.98\linewidth]{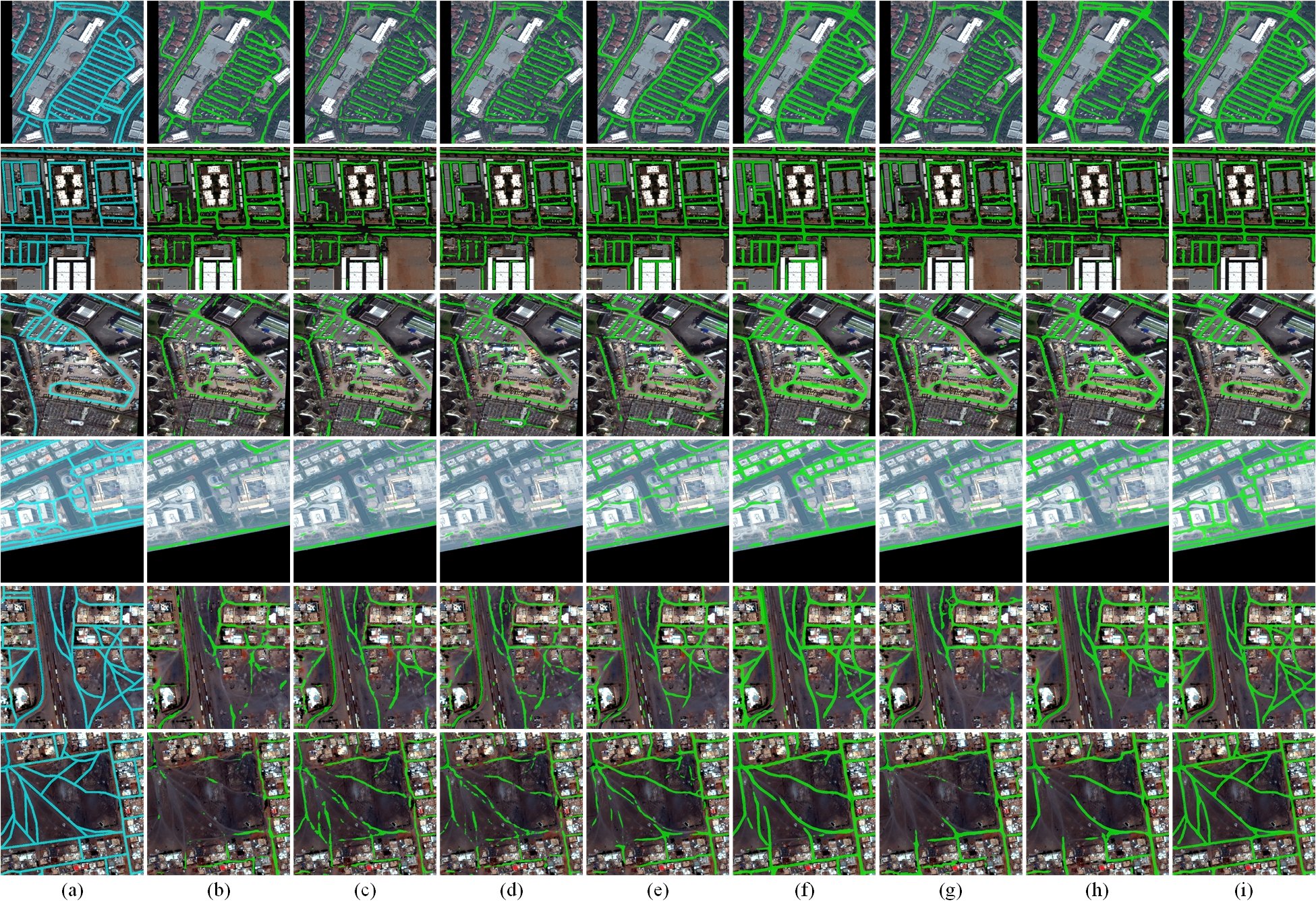}
  \vspace{-0.35cm}
   \caption{Visual results on the test images of the SpaceNet roads. Ground truth is overlaid on the orignal images in (a), followed by segmentations generated by (b) DANet, (c) UNet++, (d) RecurrUNet, (e) DSCNet, (f) SegFormer, (g) RoadFormer, (h) UCTransNet, and (i) URoadNet.}
  \label{fig9}
  \vspace{-0.15cm}
\end{figure*}

\setlength{\tabcolsep}{1.65pt}
\begin{table*}[!t]
\renewcommand\arraystretch{1.2}
  \centering
  \small
  \caption{Quantitative Results of Different Methods on the Test Images from Three Datasets at Same Resolutions ($512 \times 512$ Pixels). Our Approach Achieves the State-of-the-Art Performance, Which Outperforms the Runner-up on Each Dataset, i.e., UCTransNet by 5.74 in Road IoU on the Massachusetts, SegFormer by 0.35 in Road IoU on the DeepGlobe, and UCTransNet by 10.04 in Road IoU on the SpaceNet. The Top Results are Highlighted in {\textbf{\color[rgb]{0.4,0.4,1} blue}} and Second-Best Results in {\textbf{\color[rgb]{0.6,0.8,0.2} green}}.}
  \vspace{-0.15cm}
  \begin{tabular}{l|ccccc|ccccc|ccccc}
  \toprule[1pt]
  \multirow{2}*{\bf Architecture} & \multicolumn{5}{c}{\texttt{Massachusetts roads}} & \multicolumn{5}{|c}{\texttt{DeepGlobe roads}} & \multicolumn{5}{|c}{\texttt{SpaceNet roads}} \\
  \cmidrule(r){2-6} \cmidrule(r){7-11} \cmidrule(r){12-16}
        & {\bf OA (\scalebox{0.8}{\%})} & {\bf P (\scalebox{0.8}{\%})} & {\bf R (\scalebox{0.8}{\%})} & {\bf F1 (\scalebox{0.8}{\%})}
        & {\bf IoU (\scalebox{0.8}{\%})}
        & {\bf OA (\scalebox{0.8}{\%})} & {\bf P (\scalebox{0.8}{\%})} & {\bf R (\scalebox{0.8}{\%})} & {\bf F1 (\scalebox{0.8}{\%})}
        & {\bf IoU (\scalebox{0.8}{\%})}
        & {\bf OA (\scalebox{0.8}{\%})} & {\bf P (\scalebox{0.8}{\%})} & {\bf R (\scalebox{0.8}{\%})} & {\bf F1 (\scalebox{0.8}{\%})}
        & {\bf IoU (\scalebox{0.8}{\%})} \\
  \midrule[1pt]
  DANet\cite{c19} & 91.65 & 42.69 & 61.70 & 50.32 & 33.62 & 96.21 & 67.91 & 58.03 & 61.39 & 44.82 & 93.60 & 48.48 & 71.52 & 57.39 & 40.41 \\
  UNet++\cite{c7}  & 95.14 & 58.61 & 72.46 & 64.42 & 47.74 & 96.77 & 70.93 & 63.65 & 65.82 & 49.82 & {\bf \color[rgb]{0.6,0.8,0.2} 94.95} & 57.15 & 69.39 & 61.90 & 45.17 \\
  RecurrUNet\cite{c10} & 95.40 & 61.25 & 69.86 & 64.85 & 48.25 & 96.81 & 71.17 & 64.53 & 66.39 & 50.47 & 94.84 & 57.01 & 70.28 & 61.96 & 45.22 \\
  DSCNet\cite{c18} & 96.54 & {\bf \color[rgb]{0.4,0.4,1} 81.64} & 52.21 & 63.22 & 46.65 & 97.36 & 76.08 & 66.47 & 69.65 & 54.27 & 94.11 & {\bf \color[rgb]{0.6,0.8,0.2} 72.76} & 66.43 & 68.89 & 53.38 \\
  SegFormer\cite{c20} & {\bf \color[rgb]{0.4,0.4,1} 97.39} & 77.41 & 72.98 & 74.78 & 60.09 & {\bf \color[rgb]{0.4,0.4,1} 98.19} & {\bf \color[rgb]{0.6,0.8,0.2} 79.52} & {\bf \color[rgb]{0.6,0.8,0.2} 81.55} & {\bf \color[rgb]{0.6,0.8,0.2} 79.74} & {\bf \color[rgb]{0.6,0.8,0.2} 67.42} & 94.35 & 67.59 & {\bf \color[rgb]{0.4,0.4,1} 83.26} & {\bf \color[rgb]{0.6,0.8,0.2} 74.07} & {\bf \color[rgb]{0.6,0.8,0.2} 59.64} \\
  RoadFormer\cite{c13} & 96.02 & 62.07 & 77.47 & 69.56 & 53.82 & 97.21 & 70.39 & {\bf \color[rgb]{0.4,0.4,1} 81.65} & 73.97 & 59.63 & 93.83 & 72.00 & 65.29 & 67.49 & 51.74 \\
  UCTransNet\cite{c14} & {\bf \color[rgb]{0.6,0.8,0.2} 97.18} & 74.24 & {\bf \color[rgb]{0.6,0.8,0.2} 79.21} & {\bf \color[rgb]{0.6,0.8,0.2} 75.97} & {\bf \color[rgb]{0.6,0.8,0.2} 61.87} & 97.10 & 71.67 & 76.48 & 72.12 & 57.46 & 93.81 & 65.84 & 71.61 & 68.88 & 53.25 \\
  \midrule
  {\bf URoadNet} & 97.06 & {\bf \color[rgb]{0.6,0.8,0.2} 79.49} & {\bf \color[rgb]{0.4,0.4,1} 82.98} & {\bf \color[rgb]{0.4,0.4,1} 80.59} & {\bf \color[rgb]{0.4,0.4,1} 67.61} & {\bf \color[rgb]{0.6,0.8,0.2} 97.53} & {\bf \color[rgb]{0.4,0.4,1} 83.50} & 79.29 & {\bf \color[rgb]{0.4,0.4,1} 80.53} & {\bf \color[rgb]{0.4,0.4,1} 67.77} & {\bf \color[rgb]{0.4,0.4,1} 96.36} & {\bf \color[rgb]{0.4,0.4,1} 80.75} & {\bf \color[rgb]{0.6,0.8,0.2} 82.42} & {\bf \color[rgb]{0.4,0.4,1} 81.19} & {\bf \color[rgb]{0.4,0.4,1} 69.68} \\
  \bottomrule
  \end{tabular}
  \vspace{-0.25cm}
  \label{tab4}
\end{table*}

\subsubsection{Label Rate}
From a different perspective, we also conduct a series of ablation studies on the label rates and compare models with different settings. The results in Table~\ref{tab3} show that URoadNet achieves consistently higher performance with less label rates when comparing the representative extensions. This might be because Dual-SA has an interleaved update mechanism of multiscale features and demonstrates such embedding doesn't harm the generalization ability of U-Net architecture too much. Note that beyond a certain number of unavailable labels, our results will stop getting significantly improved, but are still on par with those of other Transformer embeddings with 100\% label rate. This further justifies the effectiveness of URoadNet.

\subsection{Visual Inspection}
We compare URoadNet against four of the most powerful U-Net extensions: UNet++\cite{c7}, RecurrUNet\cite{c10}, RoadFormer\cite{c13}, and UCTransNet\cite{c14}, two of multiscale attention methods: DANet\cite{c19} and SegFormer\cite{c20}, and one very recent strip convolution-based DSCNet\cite{c18}. Fig.~\ref{fig7}, Fig.~\ref{fig8}, and Fig.~\ref{fig9} substantiate the superiority of URoadNet in the cases of complex urban, suburban, and rural areas with unpaved, paved, and dirt roads. There, CNN variants (e.g., UNet++ and RecurrUNet) do worst because their encoders overrespond to cleaner geometric structures such as uncovered roads or paths, leading to fragmented segmentations. Attention-based methods can be taught to discount them with connectivity encoding (e.g., RoadFormer) or topology attention (e.g., UCTransNet), and in most cases, multiscale (e.g., SegFormer) does better than single-scale, but still has low saliency or semantic loss. Multiscale is also competitive on the images covering multiple scenarios, however, limited by independent free learning, resulting in incomplete prediction of road networks. Intra-road differences and inter-class similarities (such as might be caused by the shadow interference of vegetation or buildings) are also very common and challenging ``things'' in road images. From Fig.~\ref{fig7}, Fig.~\ref{fig8}, and Fig.~\ref{fig9}, we can observe several interesting behaviors:

\begin{itemize}
\renewcommand{\labelitemi}{$\vcenter{\hbox{\tiny$\bullet$}}$}
\item For the case of intra-road differences, DSCNet improves the results by replacing the square convolutions with the deformable convolutions, but the connectivity preserving is still not promising.
\item RoadFormer with deformable attention outperforms DSCNet. This improvement may be attributed to the performance boost by modeling the element relation. However, RoadFormer is unable to handle inter-class similarities, and therefore it is prone to false-positive predictions.
\item As for UCTransNet, although the channel-wise self-attention module makes it to be the best U-Net extension to tackle inter-class similarities, image-level tokens hinder local interactions across road elements.
\end{itemize}

By contrast, clear advantages of URoadNet are its immunity to the road appearance diversity, ability to deal with variations on road geometry, topology, and surrounding cover, and high prediction accuracy for the correct number of roads. For the case of intra-road differences and inter-class similarities, URoadNet also exhibits robustness.

\begin{figure*}[!t]
  \centering
  \includegraphics[width=0.98\linewidth]{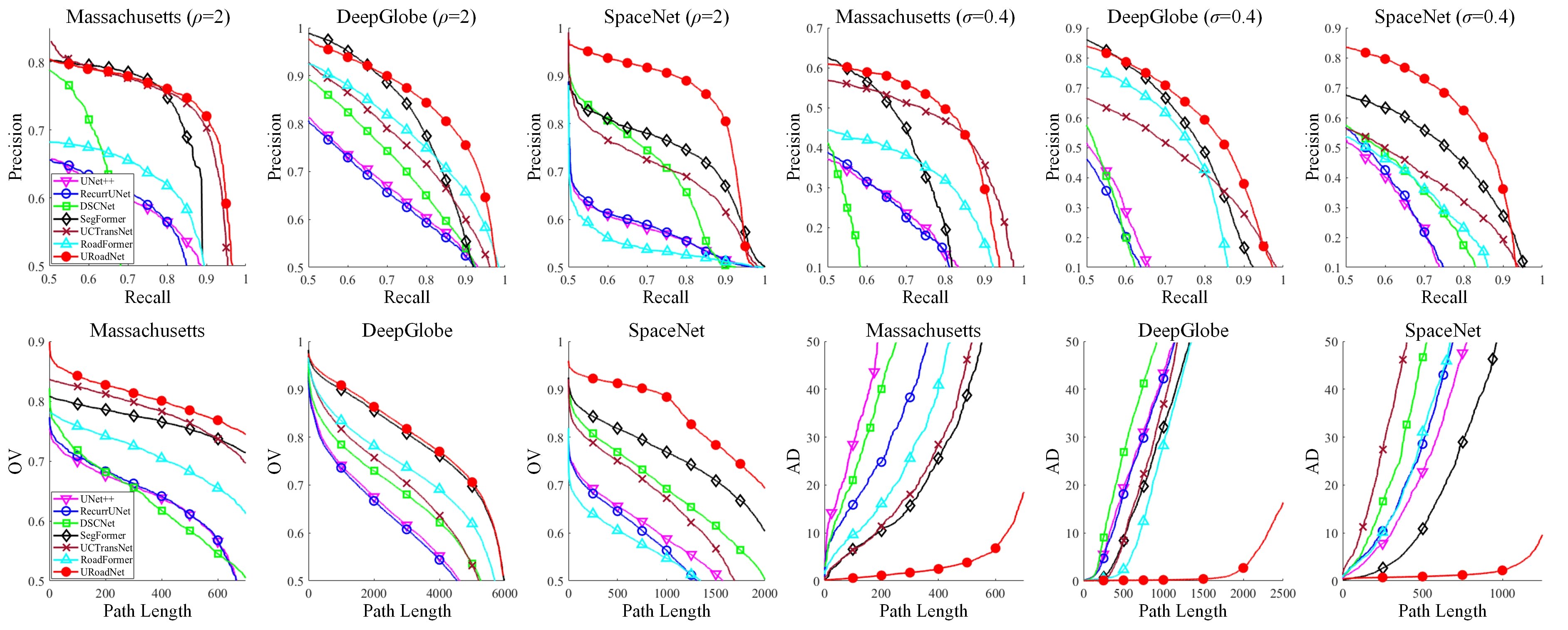}
  \vspace{-0.15cm}
   \caption{Local pixel-wise and global tracing evaluation curves on the benchmark roads datasets. Proposed URoadNet outperforms the SOTA approaches on all benchmark roads datasets and is more robust when used to trace the road structures. Note, OV is the fraction of points on the ground truth path marked as true positives, the larger the better.}
  \label{fig10}
\end{figure*}

\begin{figure*}[!t]
  \centering
  \includegraphics[width=0.95\linewidth]{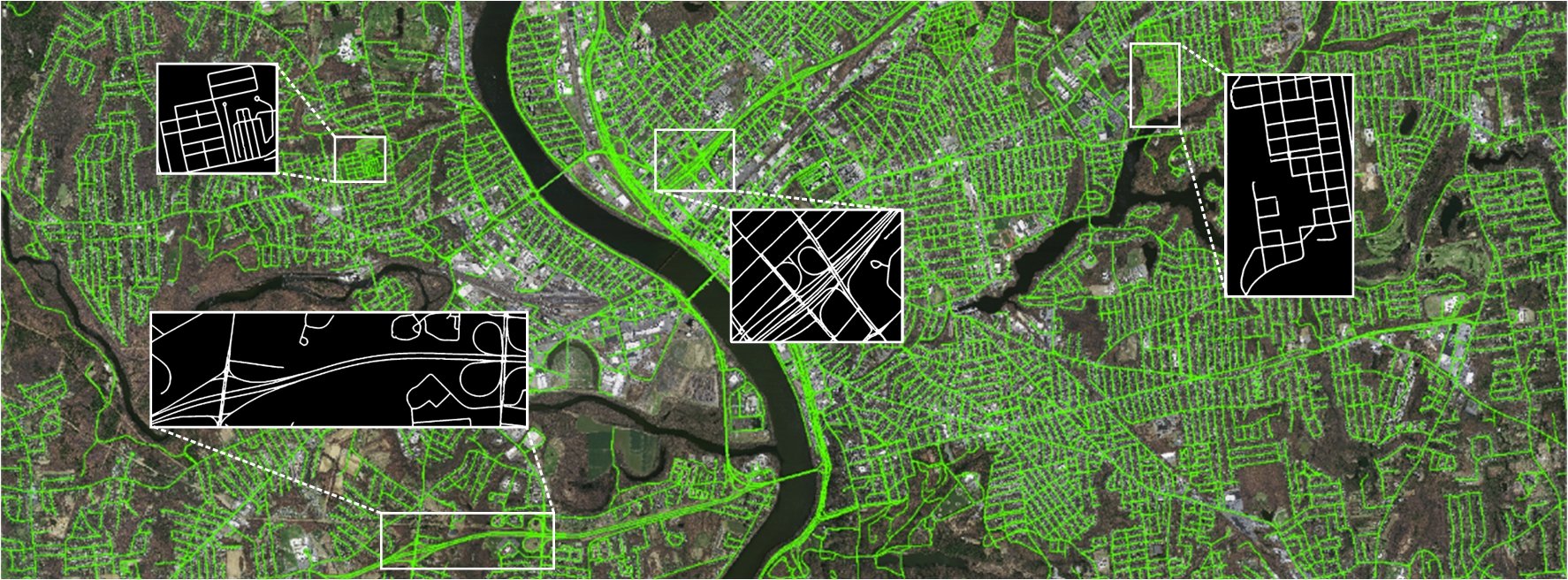}
   \caption{Visual prediction results of Large-Scale Massachusetts{\color{red}$^{\rm \bf{LS}}$} obtained by the URoadNet. More test results are presented in the supplementary materials.}
  \vspace{-0.15cm}
  \label{fig11}
\end{figure*}

\begin{figure*}[!t]
  \centering
  \includegraphics[width=0.98\linewidth]{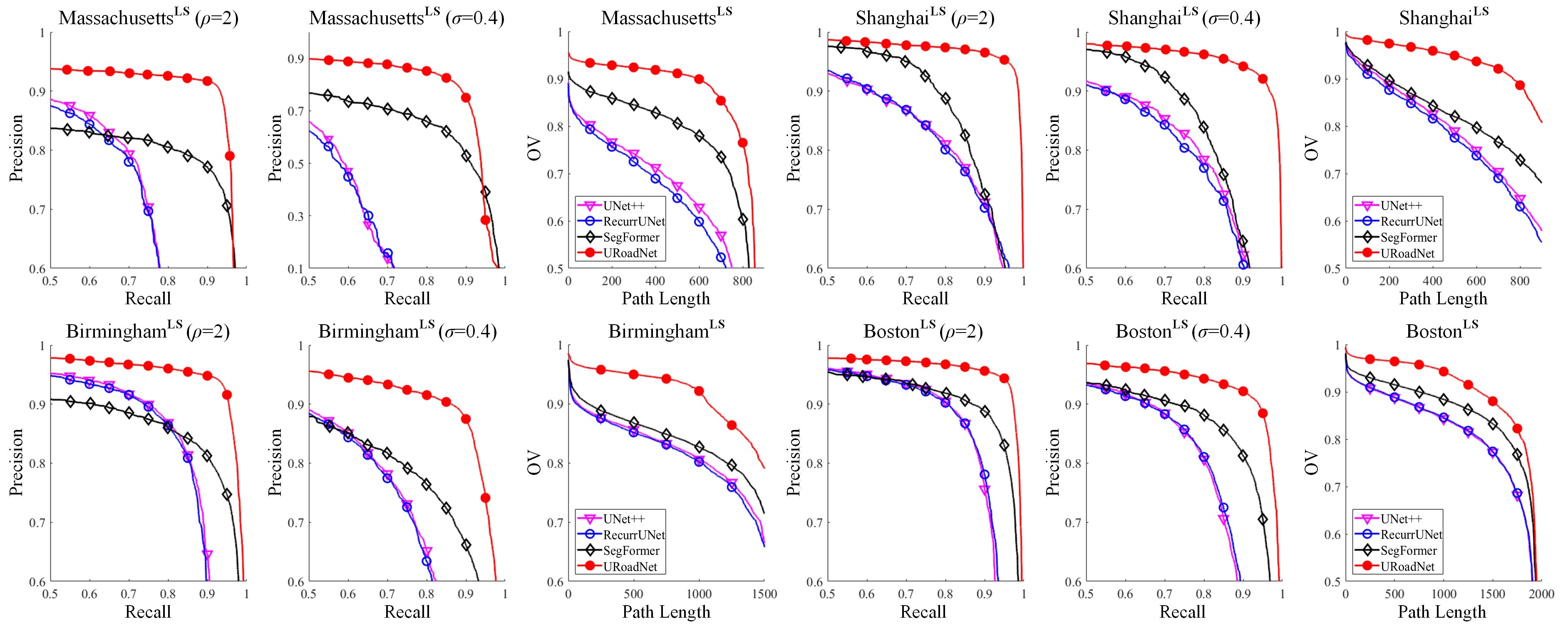}
   \caption{\footnotesize{Segmentation performance curves of precision vs. recall and tracing overlap (OV) vs. path length on Large-Scale remote sensing images.}}
  \label{fig12}
  \vspace{-0.25cm}
\end{figure*}

\subsection{Pixel-Wise Evaluation}
With the training, ablations, and testing procedure discussed above, we first use a pixel-wise evaluation to highlight significant performance improvements. The results are detailed in Table~\ref{tab4}. Our proposed URoadNet with Dual-SA embedding apparently achieves the best performances in terms of F1 and Road IoU metrics under same settings. In comparison to vanilla attention and U-Net variants, URoadNet maintains 24.40/16.72/16.37/13.52/4.57/10.43/8.45 superiority in F1 and 28.74/20.78/20.37/16.92/5.97/13.29/10.83 superiority in Road IoU over DANet/UNet++/RecurrUNet/DSCNet/SegFormer/ RoadFormer/UCTransNet, respectively, demonstrating its unique architectural superiority of enhancing attention learning in parallel by coupling local connectivity attention and global integrality attention. For precision and recall, the difference is smaller but our method still performs comprehensively better. This is a consequence of the rigid nature of each variant, which contains only attentions with a simple or independent learning.

To better quantitatively examine the quality of learnt road network via our URoadNet, we introduce a tolerance factor $\rho$ to plot precision-recall curves. A predicted centerline point is considered a true positive if it is at most $\rho$ distant from a ground truth centerline point. Fig.~\ref{fig10} top row shows the results for $\rho \!=\! 2$. We see that for a given dataset of roads and same backbone of U-Net, Dual-SA embedding can refine the road network from both local and global perspectives. Moreover, performing interleaved update further improves the modeling capacity for all the studied cases, confirming the importance of multiscale information interaction to tackle the problem. Also, since the annotations might be questionable, we introduce in segmentation case a tolerance factor $\sigma$ and eliminate from comparison pixels that are closer than $\sigma r$ from the region of a ground truth road of radius $r$. Fig.~\ref{fig10} top row shows the results for $\sigma \!=\! 0.4$. As seen, URoadNet with Dual-SA still performs better. This again confirms that the capacity of learnt roads by our URoadNet are stronger.

\subsection{Tracing Evaluation}
For each dataset, here we use the road scores predicted with URoadNet and the baselines to generate paths, randomly sample and add a fixed number of paths from the ground truth, and finally compute the OV and AD values as a function of the path length. Fig.~\ref{fig10} bottom row shows how well these paths match the ground truth path and the detailed segmentation changes. Note that the resulting paths follow the true road structures over longer distances, without being greatly disturbed by adjacent structures or background objects. In contrast with state-of-the-art, the accuracy of our approach remains higher for large values of the path length and also competitive on its small values, while the performance of the other methods decrease. All of these properties account for lower missing errors.

\subsection{Experiments on the Large-Scale Remote Sensing Images}
Another benefit of URoadNet with Dual-SA embedding is computational feasibility for Large-Scale road network prediction and reconstruction. To test the generalization ability of our algorithm, we conduct experiments on the Large-Scale areas from Massachusetts{\color{red}$^{\rm \bf{LS}}$} (Ma{\color{red}$^{\rm \bf{LS}}$}), Boston{\color{red}$^{\rm \bf{LS}}$} (Bos{\color{red}$^{\rm \bf{LS}}$}), Birmingham{\color{red}$^{\rm \bf{LS}}$} (Bhm{\color{red}$^{\rm \bf{LS}}$}), and Shanghai{\color{red}$^{\rm \bf{LS}}$} (Shh{\color{red}$^{\rm \bf{LS}}$}), covering a wide range of scenarios. The data are available at https://github.com/ wycloveinfall/large\_scale\_images and a visualization example of our result on the Massachusetts{\color{red}$^{\rm \bf{LS}}$} image is given in Fig.~\ref{fig11}. For the experiments on these Large-Scale images, we adopt the model trained on the subset of Massachusetts roads dataset and test it on the Massachusetts{\color{red}$^{\rm \bf{LS}}$}, Boston{\color{red}$^{\rm \bf{LS}}$}, Birmingham{\color{red}$^{\rm \bf{LS}}$}, and Shanghai{\color{red}$^{\rm \bf{LS}}$}, respectively. More qualitative results are exhibited in the supplementary material, indicating the performances of URoadNet.

\setlength{\tabcolsep}{0.8pt}
\begin{table}
\renewcommand\arraystretch{1.2}
  \centering
  \footnotesize
  \caption{Comparision of URoadNet with CNN Variant, RNN Variant, and Self-Attention Embedding of U-Net on Large-Scale Test Sets.}
  \begin{tabular}{l|cc|cc|cc|cc}
  \toprule[1pt]
  \multirow{2}*{\bf Architecture} & \multicolumn{2}{c}{\small{\texttt{Ma{\color{red}$^{\rm \bf{LS}}$}}}}
  & \multicolumn{2}{|c}{\small{\texttt{Bos{\color{red}$^{\rm \bf{LS}}$}}}} & \multicolumn{2}{|c}{\small{\texttt{Bhm{\color{red}$^{\rm \bf{LS}}$}}}}
  & \multicolumn{2}{|c}{\small{\texttt{Shh{\color{red}$^{\rm \bf{LS}}$}}}} \\
  \cmidrule(r){2-3} \cmidrule(r){4-5} \cmidrule(r){6-7} \cmidrule(r){8-9}
        & {\bf F1(\scalebox{0.8}{\%})} & {\bf IoU(\scalebox{0.8}{\%})} & {\bf F1(\scalebox{0.8}{\%})} & {\bf IoU(\scalebox{0.8}{\%})}
        & {\bf F1(\scalebox{0.8}{\%})} & {\bf IoU(\scalebox{0.8}{\%})} & {\bf F1(\scalebox{0.8}{\%})} & {\bf IoU(\scalebox{0.8}{\%})} \\
  \midrule[1pt]
  UNet++\cite{c7} & 66.01 & 50.97 & 80.62 & 69.28 & 77.06 & 64.84 & 75.02 & 62.37 \\
  RecurrUNet\cite{c10} & 69.40 & 53.81 & 80.65 & 69.40 & 77.77 & 65.66 & 76.91 & 63.95 \\
  SegFormer\cite{c20} & 77.27 & 65.32 & 82.05 & 72.37 & 80.03 & 68.79 & 77.19 & 65.03 \\
  \midrule
  URoadNet & {\bf 80.33} & {\bf 67.76} & {\bf 86.59} & {\bf 77.03} & {\bf 82.62} & {\bf 71.02} & {\bf 80.88} & {\bf 68.98} \\
  \bottomrule
  \end{tabular}
  \vspace{-0.25cm}
  \label{tab5}
\end{table}

After visual inspection, we report the prediction performance of three strong baselines and URoadNet. These baselines, CNN variant, RNN variant, and self-attention embedding of U-Net are trained on Massachusetts roads and tested on Ma{\color{red}$^{\rm \bf{LS}}$} [1 m/pixel], Bos{\color{red}$^{\rm \bf{LS}}$} [0.44 m/pixel], Bhm{\color{red}$^{\rm \bf{LS}}$} [0.36 m/pixel], and Shh{\color{red}$^{\rm \bf{LS}}$} [0.51 m/pixel], respectively. Table~\ref{tab5} lists the average values of local pixel-wise metrics and Fig.~\ref{fig12} provide in details the global tracing curves. They show that the accuracy of URoadNet remains higher at different levels of remote sensing resolution, and its improvements highlight the extraction of connected road network with different road types (e.g., residential), surface types (paved or unpaved), and number of lanes under complex scenarios and heavy occlusion.
%

\section{Conclusion}
In this article, we illustrate the design considerations and methods for proposing URoadNet, a novel alternative to the U-Net architecture for multiscale road network prediction which novelly models intra- and inter-road interactions in two interactive attentions: the connectivity self-attention for learning fine-grained pixel-level details and the integrality self-attention that extracts holistic global topological semantics. By enhancing attention learning in parallel through coupled connectivity and integrality self-attentions via interleaved token update, URoadNet achieves consistently superior performance over state-of-the-art on Massachusetts, DeepGlobe, SpaceNet, and Large-Scale challenges. In the future, we would expect comparable performance, particularly for the case of more general application to the field of remote sensing interpretation and image processing.


%
%
%
%
%
%

\end{document}